\documentclass[final]{cvpr}

\usepackage{times}
\usepackage{epsfig}
\usepackage{graphicx}
\usepackage{amsmath}
\usepackage{amssymb}

\usepackage{algorithm2e}

\usepackage[pagebackref=true,breaklinks=true,colorlinks,bookmarks=false]{hyperref}

\def\cvprPaperID{****} % *** Enter the CVPR Paper ID here
\def\confYear{CVPR 2021}

\begin{document}

%%%%%%%%% TITLE
\title{Motion Vector Extrapolation for Video Object Detection}

\author{Julian True\\
Ryerson University, Toronto ON\\
{\tt\small jtrue@ryerson.ca}
\and
Naimul Khan\\
Ryerson University, Toronto ON\\
{\tt\small n77khan@ryerson.ca}
}

\maketitle

%%%%%%%%% ABSTRACT
\begin{abstract}
Despite the continued successes of computationally efficient deep neural network architectures for video object detection, performance continually arrives at the great trilemma of speed versus accuracy versus computational resources (pick two). Current attempts to exploit temporal information in video data to overcome this trilemma are bottlenecked by the state-of-the-art in object detection models. We present, a technique which performs video object detection through the use of off-the-shelf object detectors alongside existing optical flow based motion estimation techniques in parallel. Through a set of experiments on the benchmark MOT20 dataset, we demonstrate that our approach significantly reduces the baseline latency of any given object detector without sacrificing any accuracy. Further latency reduction, up to 25$\times$ lower than the original latency, can be achieved with minimal accuracy loss. MOVEX enables low latency video object detection on common CPU based systems, thus allowing for high performance video object detection beyond the domain of GPU computing. The code is available at \href{https://github.com/juliantrue/movex}{https://github.com/juliantrue/movex}.
\end{abstract}

\section{Introduction}

%\begin{figure}
%\begin{center}
%\includegraphics[width=\linewidth]{imgs/comp_diagram}
%\end{center}
%   \caption{MOVEX implementation diagram. }
%\label{fig:comp_diagram}
%\end{figure}

Object detection has seen significant progress over the last several years \cite{yolov4} \cite{frcnn}. Each new iteration or approach promises higher accuracy at the cost of higher inference latency, or lower latency at the cost of lower accuracy when compared with high latency counterparts. Computing companies are in an arms race to provide hardware that offers the capability to use high accuracy models at low latencies, but this introduces, at best, a dependency on off-the-shelf GPUs and at worst a dependency on expensive niche co-processors. Despite the progress made, the trilemma persists; there is not a silver bullet to address the three constraints of accuracy, latency, and cost simultaneously.

% Optical flow methods
Optical flow has also had recent performance gains through the use of CNN architectures \cite{flownet2} \cite{flownet}. Through these methods and GPU hardware acceleration, dense optical flow techniques have become faster and more accurate over the last several years.

% Movement and video
It is well established that image content varies slowly in video data as following the same object through time can be viewed as a different task all together compared to finding unique objects every frame. Though there have been attempts to exploit this temporal information redundancy through feature propagation based on optical flow methods in the past, performance remains bottle-necked by the latency associated with a CNN inference \cite{deep_feature_flow}.

% Movex
This work proposes MOVEX, an online, real-time, method of object detection in video. Through the combination of an arbitrary off-the-shelf object detection deep neural network (DNN) with a coarse approximation of optical flow and an optimistic sparse detection propagation parallelism strategy, we demonstrate that fast, accurate, and computationally inexpensive video object detection can be achieved. Furthermore, this work demonstrates the capability to accelerate object detectors up to 25 times the original performance with minimal ($<0.01$AP) accuracy degradation. Since MOVEX does not require the use of a GPU, it enables models typically limited to the realm of GPU computing to be used on commodity CPU hardware with lower inference latency than found in GPU implementations of the same model.

% CHECKED: GOOD
\section{Related Work}

This work builds on the problem formulation provided by Zhu \textit{et al.} known as Deep Feature Flow. It frames video object detection as a two-step algorithm consisting of expensive feature extraction at sparse key-frames and feature propagation for non-key frames  through the use of a function they coin the sparse feature propagation \cite{deep_feature_flow}. 
\begin{equation}
f_k  = \mathcal{W}(f_i, M_{i \rightarrow k}, S_{i \rightarrow k})
\label{eqn:sfpf}
\end{equation} 

Equation \ref{eqn:sfpf} provides the ground work for propagating features forward from frame $I_i$ to frame $I_k$ through the use of the sparse feature propagation function $\mathcal{W}$ which accepts as input the feature map from frame $i$, the 2D optical flow field $M_{i \rightarrow k}$, and a so-called scale field $S_{i \rightarrow k}$. The authors used a CNN based method known as FlowNet \cite{flownet} to estimate the flow $M_{i \rightarrow k}$ and adding an extra channel to estimate the scale field $S_{i \rightarrow k}$. Additionally they use a ResNet 50 and 101 network with classification layers removed to use for the feature extractor backbone \cite{resnet} and turn it into an object detector through the use of an R-FCN head network on top \cite{rfcn} \cite{deep_feature_flow}.

Through this work, Zhu \textit{et al.} demonstrated that this approach was effective at reducing average latencies associated with video object detection. This technique however, did not fully redress the high latency operation of performing an inference with a CNN. Although inferencing less on a sequence of images does lower the average, it does not eliminate the necessity for a blocking high latency inference every $k$ frames.

\section{Motion Vector Extrapolation (MOVEX)}
%The MOVEX technique involves modifying the Deep Feature Flow approach to object detection in three ways.

%\begin{enumerate}
%\item Motion vectors stored as part of the video encoding were evaluated as a coarse approximation to optical flow while maintaining backward compatibility for dense optical flow methods such as FlowNet2.0 \cite{flownet}.

%\item The sparse feature propagation function was re-imagined as statistical aggregation followed by perturbation. This had the benefit of not requiring a GPU to perform pixel-wise computations for propagating features to the next frame as required by the original technique \cite{deep_feature_flow}.

%\item A parallelism strategy building on the sparse feature propagation idea was implemented to remove the bottleneck of key-frame computation present in the original sparse feature propagation approach. This method is coined optimistic sparse detection propagation.

%\end{enumerate}
\begin{figure*}
\begin{center}
\includegraphics[width=0.75\linewidth]{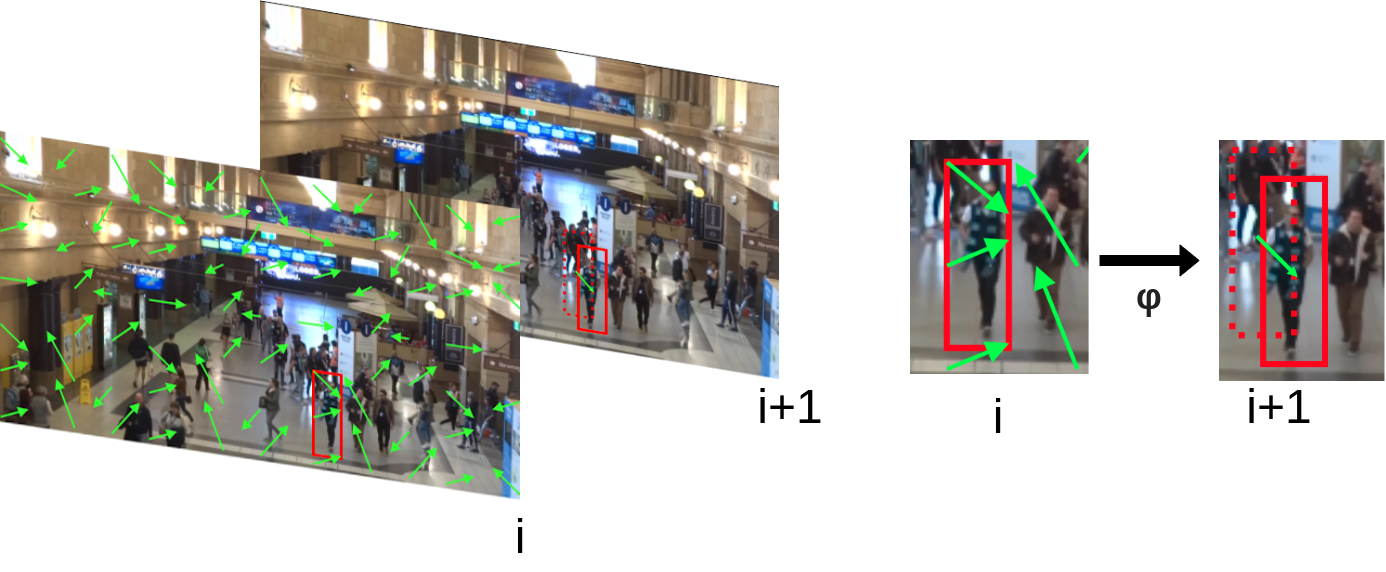}
\end{center}
   \caption{The only motion vectors considered in the source frame $i$ are those which fall in the area of the bounding box. The median perturbation of those motion vectors is computed and applied to the source bounding box in order to predict the bounding box in frame $i+1$.}
\label{fig:the_diagram}
\end{figure*}

\subsection{Coarse Optimal Flow Approximation}
Modern video codecs such as VP9 and H.265, as well as older codecs such as H.264 encode video with intra-frame and inter-frame coding techniques. In order to reduce the entropy between successive frames, these video codecs implement a macroblock (MB) structure that allows for pixel-patches to be translated within the frame before subsequently taking the difference between successive frames. These intra-frame translations that minimize the mean absolute difference (MAD) between image patches, and thus successive frames, are referred to as motion vectors (MVs) \cite{h264}.

%SSD_{m'}=\sum_{x \in X}\vert I(x+m',t-1)-I(x,t)\vert^2

%\begin{equation}
%SSD_{m'}=\sum_{x \in X}\vert I_{i-1}(x+m')-I_{i}(x)\vert^2
%\label{eqn:ssd}
%\end{equation} 

It is notable that these vectors do not specifically encode inter-frame object translations, despite their name, they encode the vector that minimizes the differences between successive macroblocks. However, it is often the case that they provide a reasonable approximation of such inter-frame object motion \cite{h264mvtracking}. It is important to note that not all H.264 encodings are created equally. Lower quality settings for such video codecs often yield poor motion vector representations while achieving their goal of minimizing MAD in less search time or fewer vectors, if any. As such, the scope of motion vector encoding's applicability to optical flow approximation remains limited to only higher quality encodings.

Despite downstream applications getting these artifacts/features for free since the are pre-computed at encoding time, very few applications make use of them. When taking into account the quality considerations, H.264 motion vectors allow for extremely fast optical flow approximations. 

%In addition to the inherent computational savings associated with using pre-computed optical flow vector approximations, the encoded vectors are also sparse (one vector per macroblock), allowing for further computational savings if subsequent computational steps are considered. Performing pixel-level optical flow computation becomes expensive as the image resolution increases, thus the necessity for GPU based parallelism becomes apparent even at reduced resolutions as shown in Table \ref{tab:resolutions}.
%\begin{table}
%\centering
%\begin{tabular}{|c|c|c|c|}
%\hline
%\textbf{Image Resolution} & \textbf{Dense Flow Vectors} & \textbf{16x16 MVs} \\
%\hline
%$720 \times 1280$ & $9.216*10^{5}$ & $3600$ \\
%\hline
%$1080 \times 1920$ & $2.0736*10^{6}$ & $8100$ \\
%\hline
%$3840 \times 2160$ & $8.2944*10^{6}$ & $32400$ \\
%\hline
%\end{tabular}
%\caption{Typical resolutions to associated numbers of optical flow vectors and motion vectors }
%\label{tab:resolutions}
%\end{table}

\subsection{Optimistic Sparse Detection Propagation}
% Explain the process
The objective of optimistic sparse detection propagation is to accept a prior set of detections and perturb them according to the sparse detection propagation function (SDPF) $\mathcal{W}$ which, similar to Zhu \textit{et Al.} takes a 2D flow field as input and rather than passing a set of features, our function accepts a prior set of object detection bounding-boxes $\mathcal{D}_i$. This difference allows our approach to be entirely model agnostic, and thus is not tied to one particular object detector.
\begin{equation}
\mathcal{D}_k  = \mathcal{W}(\mathcal{D}_i, M_{i \rightarrow k})
\label{eqn:osdpf}
\end{equation} 

The SDPF iterates over each detection $d_{j}$ in the set $\mathcal{D}_i$ and applies an aggregation function $\phi$ to the enclosed flow vectors $m_{uv}$, resulting in a net flow vector to perturb the detection with. Each 2D flow field is stored in a temporary buffer $\mathcal{B}$. 
\begin{equation}
d_k  = \phi(d_j, m_{uv})
\label{eqn:phi}
\end{equation} 

In practice, the aggregation function is simply the mean or median in $x$ and $y$, however more complicate aggregation functions that weight areas of the detection more than others can be considered. Figure \ref{fig:the_diagram} depicts the role of the aggregation function in propagating detections from the current frame to a consecutive frame.

The SDPF requires a starting set of detections to propagate forward through the video. This particular set is known as the prior detection set, which is an estimate provided from a key-frame inference. However, rather than evoke a computationally expensive and blocking DNN inference at a key-frame, this inference is computed in parallel. The object detector runs in parallel with another worker which simply iteratively applies an SPDF following equation \ref{eqn:osdpf} to the existing detection set at frame $i$.

Since there is no waiting for object detection to complete before proceeding, the process which iteratively applies the SDPF works several frames ahead of the object detector before receiving the computed detections. As such there is a discrepancy of several frames between the current set of detections and the returned detections from the object detector. However since the flow fields have been retained every frame in the flow vector buffer $\mathcal{B}$, the detections received from the inferencing process are propagated forward through iteratively applying the SDPF on said buffer of flow vectors, in order to update the prior. At the end of this update the detections at the current frame incorporate the computed detection information from the DNN worker. Articulated another way, when new information is returned for a frame that has already passed, the stored flow fields are used to re-propagate the new detection set forward to the current frame. The buffer is emptied in this update to allow for new flow vectors to be added. This process is outlined in the pseudocode algorithms \ref{algo:RPC} and \ref{algo:SDPFWorker} in the appendix.

% Explain some implications of the process
Since there is no scheduling for key-frame prior updates based on elapsed time or frame index, existing detections will continue to be propagated forward in time until the prior is updated with new information from the object detector. As such, the object detector latency does not directly contribute to the computation time of predicting detections at a frame $i$. However as the object detector latency increases, more frames will have passed during the elapsed computation time and thus will fill the motion vector buffer $\mathcal{B}$ to a greater capacity. As $\mathcal{B}$ fills with more frame data, the cost of a prior update becomes larger due to the number of frames for which detections need to be propagated forward to arrive back at the current frame $i$.

As the object detector latency increases, the interplay between updating detections based on image content versus updating based on flow vectors becomes apparent. New detection targets can only be detected with the object detector and thus higher latencies will ultimately determine performance in applications that have targets which enter and exit the image frame quickly.

\section{Experiments}
%\begin{figure*}
%\begin{center}
%\includegraphics[width=1\linewidth]{imgs/OD_latency_v_AP_and_avg_latency_v_OD_latency_combined}
%\end{center}
%   \caption{As the simulated object detector latency increases, the AP decreases, however the decrease only becomes apparently after increasing to $200ms$. For reference, a data point with $0ms$ was included to depict the original performance of the object detector detections provided with the MOT20 dataset. The average computation time per frame rises as the object detector latency increases as the prior update step becomes very time consuming due to the size of the buffer accumulated during the object detector inference time.}
%\label{fig:od_latency_v_ap_and_avg_latency}
%\end{figure*}
%\subsection{Setup}
%% Evaulation Criteria
In order to evaluate the capabilities of this approach, two critical metrics were considered: average-precision (AP) and detection latency. The dataset used to evaluate these metrics was the MOT20 dataset \cite{mot20}. The reason for using this dataset is because the data in this case is taken directly from video and maintains the temporal context between images. 

All evaluations were conducted on an Intel i7-8700K CPU 3.70GHz with Nvidia GTX 1080Ti 12GB. Tests marked with "CPU" were evaluated solely on the CPU without exposing GPU capabilities to the test, otherwise the GPU was used to evaluate.

% Experiments
%\subsection{Evaluations}
%There are four key areas to evaluate for this technique. First, the latency decrease versus accuracy for the default detections that MOT20 provides \cite{mot20}. Second, the performance implications of using H.264 motion vectors as opposed to a state-of-the-art method such as FlowNet2 \cite{flownet2}. Third, the viability of increasing accuracy while decreasing latency through the use of higher input resolution models. Fourth, the impacts of running a model on a CPU instead of a GPU. These four areas will be evaluated through AB testing a model with and without the attribute of interest. 

%\section{Results}
\begin{table*}
\centering
\begin{tabular}{|l|c|c|}
\hline
\textbf{Method} &  \textbf{Avg Latency (\textit{ms}) $\downarrow$} & \textbf{AP $\uparrow$} \\
\hline
FRCNN \cite{mot20} & $ 131.52 $ & $ 63.0 $ \\
\hline
FRCNN \cite{mot20} w/ MOVEX + FlowNet2 &  $ 263.25 $ & $ 62.4 $ \\
\hline
FRCNN \cite{mot20} w/ MOVEX + H.264 MVs & $ 12.79 $  & $ 62.3 $ \\
\hline
YOLOv4 \cite{yolov4} (960x960) &  $ 79.46 $ & $40.2$ \\
\hline
YOLOv4 \cite{yolov4} (960x960) w/ MOVEX + H.264 MVs &  $ 11.06 $ & $39.8$ \\
\hline
YOLOv4 \cite{yolov4} (416x416) &  $ 26.34 $ & $ 26.1 $ \\
\hline
YOLOv4 \cite{yolov4} (416x416) on CPU &  $ 190.41 $ & $ 26.1 $ \\
\hline
YOLOv4 \cite{yolov4} (416x416) on CPU w/ MOVEX + H.264 MVs &  $ 7.53 $ & $25.2$ \\
\hline
\end{tabular}
\caption{Evaluation of MOVEX with H.264 MVs or FlowNet2 optical flow against baseline Faster R-CNN model used for MOT20 public detections without augmentation \cite{mot20}. As expected, FlowNet2 flow vectors are more accurate than the approximation provided by the H.264 motion vectors, however this better accuracy comes at a cost of high inference latency. Varying the input resolutions of two YOLOv4 models \cite{yolov4} trained on the COCO dataset \cite{coco} demonstrates the accuracy gains possible without sacrificing inference latency.}
\label{tab:comparison}
\end{table*}

\subsection{H.264 Motion Vectors and FlowNet2.0}
Examining the performance results in Table \ref{tab:comparison}, the latency of the original Faster R-CNN model is reduced by a factor of $10.3\times$ when run with MOVEX using the H.264 MVs. However, when using FlowNet2 as the source for the motion vectors, performance is greatly impacted, presenting a $2.0\times$ increase in latency. The FlowNet2-s model was used to compute the optical flow which claimed to have a runtime of approximately $7ms$ on a GTX 1080Ti \cite{flownet2}, however performance when running it was no where near this as model forward passes were routinely reaching $100ms$. The AP differences between the three evaluated Faster R-CNN models demonstrates that the use of MOVEX decreases the AP of the baseline model by approximately $0.007$ AP when using H.264 MVs but only a decrease of $0.006$ AP when using the FlowNet2.0 model.

%One reason for this discrepancy may be due the increased image resolutions of the images in the MOT20 dataset compared to the images the authors demonstrate on originally. Images in MOT20 are all $1920 \times 1080$ while images in KITTI 2015 are $1248 \times 384$ \cite{kittiart} \cite{kitticonf} and the Sintel dataset images are $1024 \times 436$ \cite{sintel}. This resolution difference results in approximately a $4\times$ decrease in image pixels when compared with the MOT20 data, and thus, could have lead to the latency increases observed. 
\subsection{Hi-Resolution Versus Low-Resolution}
The effect of increasing input resolution for CNN object detectors is known to increase their accuracy. Shown in Table \ref{tab:comparison}, YOLOv4 trained on the COCO dataset \cite{coco} is compared against itself at two different resolutions $416\times416$ versus $960\times960$, resulting in APs of $0.261$ and $0.402$ respectively. This confirms the relationship between input resolution and accuracy, but also demonstrates an opportunity for the MOVEX augmentation. Consider that when using the higher resolution model with MOVEX, the AP drops by a mere $0.002$ yet the latency falls below that of the original low resolution model. A latency decrease of $7.18\times$ compared to the original high resolution model.

\subsection{CPU Versus GPU}
Continuing in the vein of accelerating typically high latency object detectors, consider CPU versus GPU object detection latency. It is well known by practitioners that hardware accelerators such as GPUs or TPUs are needed to achieve low latency computation with CNNs. This point is further articulated by the latency data point given by running YOLOv4 with an input resolution of $416\times416$ on a CPU. This yields a latency of $190.41ms$, which is far too large for any real-time application. When using MOVEX in conjunction with this model however, the latency falls lower than the original GPU computation latency, resulting in a latency reduction by $25.29\times$ and falling 0.009 AP. 

Such latencies for large model such as YOLOv4 on a CPU have yet to be achieved with existing inference acceleration methodologies. GPU computational resources are orders of magnitude more expensive than standard CPU based systems. Employing MOVEX in systems looking to perform object detection on video data would lead to large cost savings by switching from GPU to CPU focused computing. Furthermore, emerging applications in edge computing where cost, space, and computing capabilities are typically limited would greatly benefit from using this technique since modern GPU centered computing often clashes directly with these constraints.

%\section{Future Work}
%There are several opportunities for further improvements on this technique. Further parallelism improvements are possible within the optimistic sparse detection propagation function since the application of the function to a detection is independent of the other detections, allowing for updating multiple detections simultaneously. This would allow for further latency reduction but would add a non-negotiable dependency on having a GPU as the number of detections could be large.

\section{Conclusion}
We presented MOVEX, a technique that can be applied to an arbitrary off-the-shelf object detector and reduce its inference latency on video data  by large margins while sacrificing minimal accuracy. We have demonstrated that MOVEX improves performance for existing object detection models, for which, online real-time video object detection would not have been possible prior. Additionally, we have shown that accuracy improvements are possible without sacrificing latency through increasing the resolution of models and using these models with MOVEX. Lastly, MOVEX allows for models typically restricted to the domain of GPU or TPU computing, due to latency concerns, to expand to less expensive CPU devices.

{\small
\bibliographystyle{ieee_fullname}
\bibliography{movexbib}
}

\section{Appendix}

\begin{algorithm}
\SetAlgoLined
\SetKwInOut{Input}{input}
\Input{$image$ of resolution $w\times h$}
\Begin{
	img $\leftarrow preprocess\_image(img)$\;
	detections $\leftarrow object\_detection(img)$\;
	return $detections$
}
\caption{Object detection remote procedure call (RPC) psuedocode.}
\label{algo:RPC}
\end{algorithm}

\begin{algorithm}
\SetKwInOut{Input}{input}
\Input{$N$ length sequence of $images$ of resolution $w\times h$}
\SetAlgoLined

\Begin{
m $\leftarrow 1$\;
$send\_img\_to\_object\_detection\_RPC(img_m)$\;
detections$_m$ $\leftarrow wait\_for\_detections$\;

initialize $MV$ buffer $\mathcal{B}$

\For{$i \leftarrow 2$ \KwTo $N$}{
	add $MV_{i \rightarrow i+1}$ to $\mathcal{B}$\;

	\eIf {$detections\_received\_from\_RPC$ }{
		prior$_m$ $\leftarrow$ RPC\_detections\;	
		
		\For{$m$ \KwTo $i$}{
			apply $MV_m$ from $\mathcal{B}_m$ to $prior_m$ using $\mathcal{W}$
		}		
		m $\leftarrow$ i\;
		empty MV buffer $\mathcal{B}$\;
		detections$_i$ $\leftarrow prior_m$\;
		$send\_img\_to\_object\_detection\_RPC(img_i)$\;
	}
	{
		detections$_i$ $\leftarrow \mathcal{W}(prior_m, MV_{m \rightarrow m+1})$\;		
	}					
}
}
\caption{SDPF worker process psuedocode.}
\label{algo:SDPFWorker}
\end{algorithm}

\end{document}